\pdfoutput=1
\documentclass[11pt]{article}

\usepackage{EMNLP2022}

\usepackage{times}
\usepackage{latexsym}

\usepackage[T1]{fontenc}
\usepackage[utf8]{inputenc}

\usepackage{microtype}

\usepackage{inconsolata}

\usepackage{graphicx}
\usepackage{covington}
\usepackage{lipsum}
\usepackage{stfloats}
\title{Collateral facilitation in humans and language models}

\author{James A. Michaelov \\
  Department of Cognitive Science \\
  University of California, San Diego \\
  \texttt{j1michae@ucsd.edu} \And
  Benjamin K. Bergen \\
  Department of Cognitive Science \\
  University of California, San Diego \\
  \texttt{bkbergen@ucsd.edu} }

\begin{document}
\maketitle
\begin{abstract}
Are the predictions of humans and language models affected by similar things? Research suggests that while comprehending language, humans make predictions about upcoming words, with more predictable words being processed more easily. However, evidence also shows that humans display a similar processing advantage for highly anomalous words when these words are semantically related to the preceding context or to the most probable continuation. Using stimuli from 3 psycholinguistic experiments, we find that this is also almost always also the case for 8 contemporary transformer language models (BERT, ALBERT, RoBERTa, XLM-R, GPT-2, GPT-Neo, GPT-J, and XGLM). We then discuss the implications of this phenomenon for our understanding of both human language comprehension and the predictions made by language models.
\end{abstract}

\section{Introduction}
Humans process words more easily when they more contextually predictable, whether predictability is determined by humans \citep{fischler_1979_AutomaticAttentionalProcesses,brothers_2021_WordPredictabilityEffects} or language models \citep{mcdonald_2003_EyeMovementsReveal,levy_2008_ExpectationbasedSyntacticComprehension,smith_2013_EffectWordPredictability}. Work on the N400, a neural signal of processing difficulty, has provided evidence that the neurocognitive system underlying human language comprehension preactivates words based on the extent to which they are predictable from the preceding context---thus, predictable words are easier to process because they or their features have already been activated before they are encountered \citep{kutas_1984_BrainPotentialsReading,vanpetten_2012_PredictionLanguageComprehension}. This has led many to argue that we should consider the human language comprehension system to be engaging in prediction \citep{delong_2005_ProbabilisticWordPreactivation,kutas_2011_LookWhatLies,vanpetten_2012_PredictionLanguageComprehension,bornkessel-schlesewsky_2019_NeurobiologicallyPlausibleModel,kuperberg_2020_TaleTwoPositivities,delong_2020_ComprehendingSurprisingSentences,brothers_2021_WordPredictabilityEffects}. 

However, words that are either semantically related to the elements of the preceding context or to the most likely next word are also processed more easily, even if they are semantically implausible and ostensibly unpredictable. These are known as \textit{related anomaly} effects. For an example of the former, consider the sentences in (\ref{ex:metusalem_example1}) that were used as experimental stimuli by \citet{metusalem_2012_GeneralizedEventKnowledge}.

\begin{example}
\label{ex:metusalem_example1}
My friend Mike went mountain biking recently. He lost control for a moment and ran right into a tree. It's a good thing he was wearing his \_\_\_\_\_\_.
\begin{itemize}
\item[(a)] \textit{helmet}
\item[(b)] \textit{dirt}
\item[(c)]  \textit{table}
\end{itemize}
\end{example}

\textit{Helmet} is the most predictable continuation of the sentence, as determined based on cloze probability \citep{taylor_1953_ClozeProcedureNew,taylor_1957_ClozeReadabilityScores}---the proportion of people to fill in a gap in a sentence with a specific word. Thus, unsurprisingly, \textit{helmet} elicited the smallest N400 response, indicating that it is most easily processed. \textit{Dirt} and \textit{table} are both implausible continuations, and equally improbable based on human responses (both have a cloze probability of zero). Yet \citet{metusalem_2012_GeneralizedEventKnowledge} found that \textit{dirt}, which is semantically related to the preceding context of \textit{mountain biking}, elicits a smaller N400 response than \textit{table}, which is not. This suggests that something about \textit{dirt}'s relation to the \textit{mountain biking} event causes it to be preactivated more than \textit{table}, despite their seemingly equal implausibility and unpredictability.

The sentences in (\ref{ex:ito_example1}), used as experimental stimuli by \citet{ito_2016_PredictingFormMeaning}, provide an example of the other previously-discussed form of related anomaly---where a word semantically related to the most probable continuation (in this case, that with the highest cloze) is easier to process than one that is not. Even though \textit{tail} and \textit{tyre} are both implausible continuations with a cloze probability of zero, \citet{ito_2016_PredictingFormMeaning} find that \textit{tail}, which is semantically-related to the highest-cloze continuation \textit{dog}, elicits a smaller N400 response than \textit{tyre}, which is not.

\begin{example}
\label{ex:ito_example1}
Meg will go to the park to walk her \_\_\_\_\_\_ tomorrow.
\begin{itemize}
\item[(a)] \textit{dog}
\item[(b)] \textit{tail}
\item[(c)]  \textit{tyre}
\end{itemize}
\end{example}

In sum, words related to elements of the preceding context or to the most probable continuation of a sequence appear to be more preactivated in the brain than words that are not, even when both are highly anomalous. This effect has been replicated many times (\citealp{kutas_1984_BrainPotentialsReading,kutas_1984_WordExpectancyEventrelated,kutas_1993_CompanyOtherWords,federmeier_1999_RoseAnyOther,metusalem_2012_GeneralizedEventKnowledge,rommers_2013_ContentsPredictionsSentence,ito_2016_PredictingFormMeaning,delong_2019_SimilarTimeCourses}; for review see \citealp{delong_2019_SimilarTimeCourses}).

The key question, therefore, is whether the same neurocognitive system underlying the predictability effects on the N400 also underlie related anomaly effects. Under one account \citep{delong_2019_SimilarTimeCourses,delong_2020_ComprehendingSurprisingSentences}, the predictive system that underlies predictability effects also leads to these related anomalous words being `collaterally facilitated' \citep[p.~1045]{delong_2020_ComprehendingSurprisingSentences} due to their shared semantic features. Under this account, therefore, related anomaly effects can all be explained as by-products of our predictive system and the semantic organization of information in the brain. However, there is no direct evidence that this is the case---in fact, given the metabolic costs of preactivation \citep{brothers_2021_WordPredictabilityEffects}, it may intuitively seem unlikely that an efficient predictive system would lead to implausible and otherwise anomalous words being preactivated. In fact, many researchers have argued that one or more associative mechanisms are required to explain related anomaly and other similar effects \citep{lau_2013_DissociatingN400Effects,ito_2016_PredictingFormMeaning,frank_2017_WordPredictabilitySemantic,federmeier_2021_ConnectingConsideringElectrophysiology}.

As systems designed specifically to predict the probability of a word given its context, language models offer a means to test the viability of the former hypothesis. If language models calculate that related but anomalous words are more predictable than unrelated anomalous words, this would demonstrate that related anomaly effects can be produced by a system engaged in prediction alone. This would show that it is possible that related anomalies can be `collaterally facilitated' \citep[p.~1045]{delong_2020_ComprehendingSurprisingSentences} by a predictive mechanism in human language comprehension. Thus, it would remove the need to posit additional associative mechanisms on the basis of related anomaly effects, which could greatly simplify our understanding of human language comprehension.

This is what we test in the present study. We run the stimuli from 3 psycholinguistic experiments carried out in English \citep{ito_2016_PredictingFormMeaning,delong_2019_SimilarTimeCourses,metusalem_2012_GeneralizedEventKnowledge} through 8 contemporary transformer language models \citep{devlin_2019_BERTPretrainingDeep,radford_2019_LanguageModelsAre,liu_2019_RoBERTaRobustlyOptimized,lan_2020_ALBERTLiteBERT,conneau_2020_UnsupervisedCrosslingualRepresentation,black_2021_GPTNeoLargeScale,wang_2021_GPTJ6BBillionParameter,lin_2021_FewshotLearningMultilingual}, calculating the surprisal (negative log-probability) of each word for which the N400 was measured. We then compare whether, in line with the N400 response, anomalous words that are semantically related to the context have significantly lower surprisals than unrelated words.

\section{Related work}
There have been a wide range of attempts to computationally model the N400 \citep{parviz_2011_UsingLanguageModels,laszlo_2012_NeurallyPlausibleParallel,laszlo_2014_PSPsERPsApplying,rabovsky_2014_SimulatingN400ERP,frank_2015_ERPResponseAmount,ettinger_2016_ModelingN400Amplitude,cheyette_2017_ModelingN400ERP,brouwer_2017_NeurocomputationalModelN400,rabovsky_2018_ModellingN400Brain,venhuizen_2019_ExpectationbasedComprehensionModeling,fitz_2019_LanguageERPsReflect,aurnhammer_2019_EvaluatingInformationtheoreticMeasures,michaelov_2020_HowWellDoes,merkx_2021_HumanSentenceProcessing,uchida_2021_ModelOnlineTemporalSpatial,szewczyk_2022_ContextbasedFacilitationSemantic,michaelov_2022_ClozeFarN400}. One of the most successful and influential approaches has been to model the N400 using the surprisal calculated from neural language models---surprisal has been found to be a significant predictor of single-trial N400 data \citep{frank_2015_ERPResponseAmount,aurnhammer_2019_EvaluatingInformationtheoreticMeasures,merkx_2021_HumanSentenceProcessing,michaelov_2021_DifferentKindsCognitive,szewczyk_2022_ContextbasedFacilitationSemantic,michaelov_2022_ClozeFarN400}, and has been found to be similar to the N400 response in how it is affected by a range of experimental manipulations \citep{michaelov_2020_HowWellDoes,michaelov_2021_DifferentKindsCognitive}. A key finding is that better-performing and more sophisticated language models perform better at predicting the N400 \citep{frank_2015_ERPResponseAmount,aurnhammer_2019_EvaluatingInformationtheoreticMeasures,michaelov_2020_HowWellDoes,merkx_2021_HumanSentenceProcessing,michaelov_2021_DifferentKindsCognitive,michaelov_2022_ClozeFarN400}. For this reason, we use contemporary transformer language models in the present study.

We use experimental stimuli from 3 experiments. Stimuli from one of these experiments \citep{ito_2016_PredictingFormMeaning} have been previously used in computational analyses of the N400. This is one of several sets that \citet{michaelov_2020_HowWellDoes} attempt to model using recurrent neural network (RNN) language models, finding that they can indeed calculate that words related to the highest-cloze continuation are more predictable than unrelated words. In the present study, we test whether this result can be replicated on a larger number of language models, and specifically, transformer language models.

There has also been work looking at how language models deal with semantic relatedness to the highest-cloze continuation based on stimuli from other N400 experiments. \citet{michaelov_2020_HowWellDoes}, for example, find that in cases where the related and unrelated words are both plausible, the related continuations are more strongly predicted by RNNs \citep{gulordava_2018_ColorlessGreenRecurrent,jozefowicz_2016_ExploringLimitsLanguage}, in line with the original N400 results \citep{kutas_1993_CompanyOtherWords}. \citet{michaelov_2021_DifferentKindsCognitive} conceptually replicate this finding on a different dataset \citep{bardolph_2018_SingleTrialEEG} using one of the same RNNs \citep{jozefowicz_2016_ExploringLimitsLanguage} and GPT-2 \citep{radford_2019_LanguageModelsAre}. However, these prior efforts differ from the present study in that they investigate N400s and surprisal to words that are all plausible continuations of the sentence, and where they both have a low but generally non-zero cloze probability. In the stimuli analyzed in the present study, by contrast, both the related and unrelated words are anomalous---they have a cloze probability of zero, and are implausible continuations. Thus, their preactivation does, at least intuitively, appear to be more clearly `collateral'.

We are only aware of one previous study that directly compares the predictions of transformers and the human N400 response on related anomaly stimuli. \citet{ettinger_2020_WhatBERTNot} evaluates BERT in terms of its similarity to cloze---because the predictions of a language model, being incremental, may show similar effects to those found in the N400 (see also \citealp{michaelov_2020_HowWellDoes} for discussion). For this reason, \citet{ettinger_2020_WhatBERTNot} tests how good BERT is at predicting the highest-cloze (most probable) continuations in the stimuli over anomalous but semantically related continuations, but does not directly look at the related anomaly effect---whether the related anomalous continuations are more strongly predicted than the unrelated anomalous continuations. Thus, to the best of our knowledge, the present study is the first to investigate whether the predictions of transformer language models display related anomaly effects like humans do.

Finally, there has been some work investigating whether language models display priming effects \citep[e.g.][]{prasad_2019_UsingPrimingUncover,misra_2020_ExploringBERTSensitivity,kassner_2020_NegatedMisprimedProbes,lin_2021_FewshotLearningMultilingual,lindborg_2021_MeaningBrainsMachines}. The effect found by \citet{metusalem_2012_GeneralizedEventKnowledge}---that words related to the events described in the context are preactivated more strongly than words that are not---is a form of semantic priming, as it results in the increased preactivation of a word based on the semantic content stimulus that has been recently encountered (i.e. the event described in the preceding linguistic context). Thus, our investigation of the patterns in the prediction of the the stimuli from \citet{metusalem_2012_GeneralizedEventKnowledge} is intended to further our knowledge of priming in language models---specifically, whether there are systematic ways in which context shapes the extent to which anomalous words are predicted.

\section{General Method}
In this study, we took the stimuli from a range of experiments \citep{ito_2016_PredictingFormMeaning,delong_2019_SimilarTimeCourses,metusalem_2012_GeneralizedEventKnowledge} and ran them through a number of transformer language models. We used the \textit{transformers} \citep{wolf_2020_TransformersStateoftheArtNatural} implementations of the (largest and most up-to-date versions of each of the) following models: BERT \citep{devlin_2019_BERTPretrainingDeep}, RoBERTa \citep{liu_2019_RoBERTaRobustlyOptimized}, ALBERT \citep{lan_2020_ALBERTLiteBERT}, XLM-R \cite{conneau_2020_UnsupervisedCrosslingualRepresentation}, GPT-2 \citep{radford_2019_LanguageModelsAre}, GPT-Neo \citep{black_2021_GPTNeoLargeScale}, GPT-J \citep{wang_2021_GPTJ6BBillionParameter}, and XGLM \citep{lin_2021_FewshotLearningMultilingual}. We chose these models to cover a number of both autoregressive (GPT-2, GPT-Neo, GPT-J, XGLM) and masked (BERT, RoBERTa, ALBERT, XLM-RoBERTa) language model architectures. Given the recent increase in popularity of multilingual language models, we also made sure to include one autoregressive (XGLM) and one masked (XLM-RoBERTa) multilingual language model, in case there is a difference based on the number of languages that a model is trained on.

All experimental stimuli used in the present study have been made available by the original authors of their respective papers as appendices or supplementary materials. In our analysis, we truncated all stimuli to be the preceding context of the critical word (the word for which the N400 was measured). We then used the language models to calculate the probability of the next word, and negative log-transformed (using a logarithm of base 2, following \citealp{futrell_2019_NeuralLanguageModels}) these probabilities to calculate the surprisal of each word. For words not present in the vocabulary of each model, we tokenized the word, and then progressively calculated the surprisal of each sub-word token given the preceding context; with the sum of all the surprisals (equivalent to the the negative log-probability of the product of all the probabilities) being used as the total surprisal for the word. In this way, we calculated the surprisal of each critical word given its preceding context only.

All graphs and statistical analyses were created and run in \textit{R} \citep{rcoreteam_2020_LanguageEnvironmentStatistical} using \textit{Rstudio} \citep{rstudioteam_2020_RStudioIntegratedDevelopment} and the \textit{tidyverse} \citep{wickham_2019_WelcomeTidyverse}, \textit{lme4} \citep{bates_2015_FittingLinearMixedeffects}, and \textit{lmerTest} \citep{kuznetsova_2017_LmerTestPackageTests} packages. All reported $p$-values are corrected for multiple comparisons based on false discovery rate across all statistical tests carried out \citep{benjamini_1995_ControllingFalseDiscovery}. Because of this correction procedure, if any models display related anomaly effects, this is evidence that prediction alone can account for them. 

All of the code for running the experiments and carrying out the statistical analyses is provided at \url{https://github.com/jmichaelov/collateral-facilitation}.

\section{Experiment 1: \citet{ito_2016_PredictingFormMeaning}}

\begin{figure*}
    \centering
    \includegraphics[width=\textwidth]{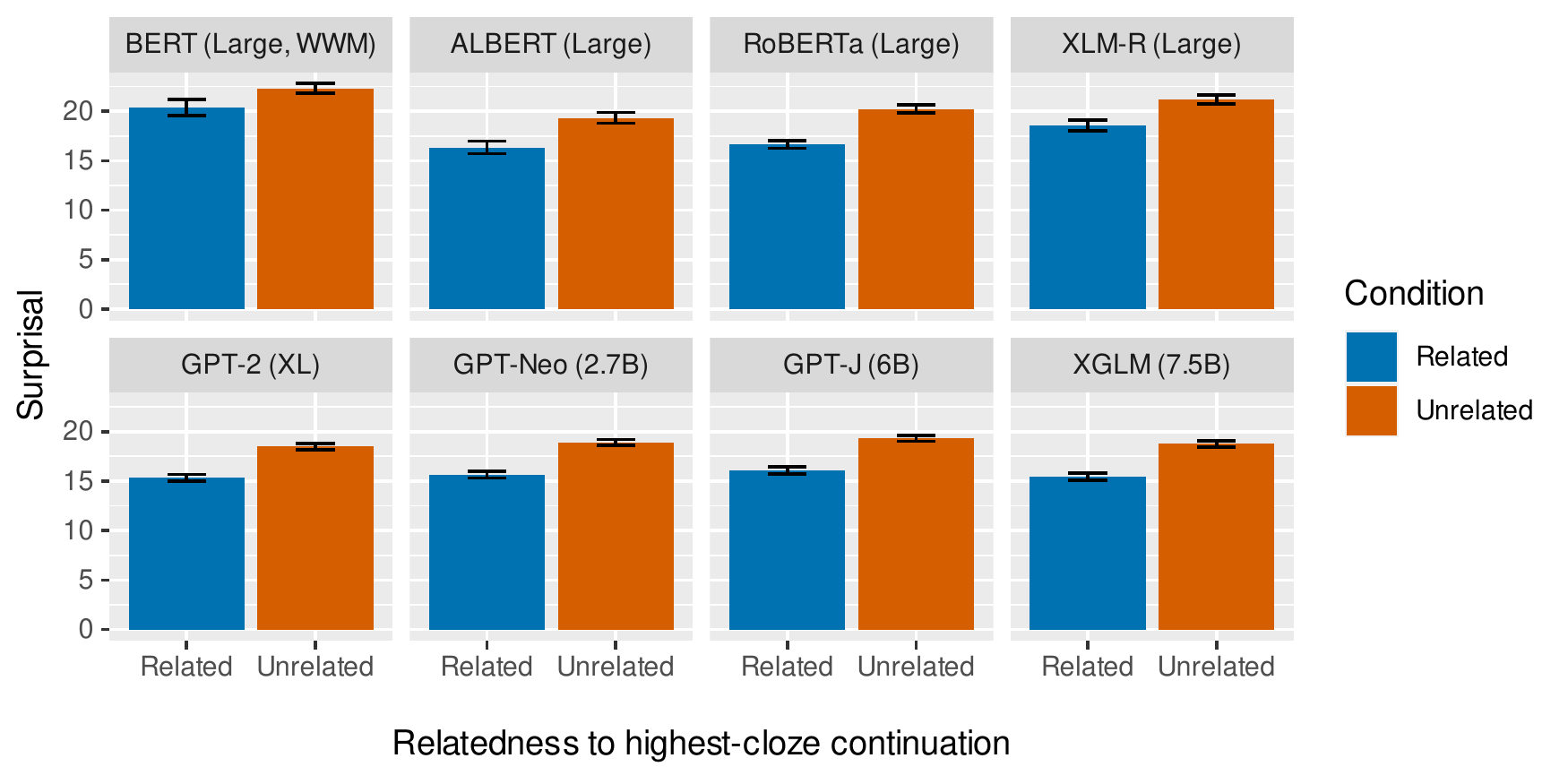}
    \caption{Mean surprisal elicited by each language model for the 
    \citet{ito_2016_PredictingFormMeaning} stimuli related and unrelated to the most probable (highest-cloze) continuation of each sentence. Error bars indicate standard error.}
    \label{fig:ito}
\end{figure*}

\subsection{Introduction}

We begin with \citet{ito_2016_PredictingFormMeaning}, who investigated whether relatedness to the highest-cloze continuation of a given sentence impacts the amplitude of the N400 response. They presented human participants with experimental stimuli that included a word that was either the highest-cloze continuation of a sentence, semantically related to that highest-cloze continuation, similar to the highest-cloze continuation in terms of their form (e.g. \textit{hook} and \textit{book}), or unrelated. For the purposes of the present study, we are interested in semantic relatedness and thus do not consider the formal relatedness condition. Thus, we look at the stimuli from the three experimental conditions exemplified in (\ref{ex:ito_example2})---an example of Predictable, Related, and Unrelated continuations for one sentence frame.

\begin{example}
\label{ex:ito_example2}
Lydia cannot eat anymore as she is so \_\_\_\_\_\_ now.
\begin{itemize}
\item \textit{full} (Predictable)
\item \textit{half} (Related)
\item \textit{mild} (Unrelated)
\end{itemize}
\end{example}

\citet{ito_2016_PredictingFormMeaning} find that related continuations elicit a smaller N400 response than unrelated continuations. As stated, this finding was successfully modeled using the surprisal of two RNN language models by \citet{michaelov_2020_HowWellDoes}. 

In the present study, we aim to investigate whether this can be replicated with contemporary transformer language models. Thus far, only one study \citep{merkx_2021_HumanSentenceProcessing} has directly compared the N400 prediction capabilities of RNNs and transformers while matching number of parameters, training data, and language modeling performance, finding that transformers are better predictors of N400 amplitude overall. We might therefore expect that the transformers used in the present study should model the related anomaly effect found by \citet{ito_2016_PredictingFormMeaning} at least as well as the RNNs used by \citet{michaelov_2020_HowWellDoes}. However, a key feature of \citeauthor{merkx_2021_HumanSentenceProcessing}'s \citeyearpar{merkx_2021_HumanSentenceProcessing} study is that it uses naturalistic stimuli. This makes the experiment more ecologically valid, but as has been pointed out \citep{michaelov_2020_HowWellDoes,brothers_2021_WordPredictabilityEffects}, this means that we cannot tell whether the higher correlation between surprisal and N400 amplitude is due to any factors that we are interested in investigating---\citet{merkx_2021_HumanSentenceProcessing} do not consider how relatedness to a previously-mentioned event or to most predictable continuation impacts surprisal and the N400. For this reason, it is in fact far from clear that we should expect this specific related anomaly effect to be modeled as well by transformers as by RNNs. However, if it is, this would demonstrate the effect in two different language model architectures, further strengthening the idea that a predictive system alone can explain related anomaly effects.

Thus, in the present study, we investigate whether the results of \citet{michaelov_2020_HowWellDoes} replicate beyond the two RNNs tested, and crucially, whether the results replicate with transformer language models. Specifically, we test whether the surprisal elicited by implausible stimuli related to the highest-cloze continuation is lower than the surprisal elicited by implausible stimuli unrelated to the highest-cloze continuation.

\begin{table}[]
\renewcommand{\arraystretch}{1.2}
    \centering
        \begin{tabular}{llr}
        \hline
        \textbf{Model}       & \textbf{Test Statistic}    & \textbf{Corrected \textit{p}}      \\ \hline
        BERT    & $F(1,120) = 7.15  $& $0.0093$           \\
        ALBERT      & $F(1,92) = 20.6  $& $<0.0001$ \\
        RoBERTa     & $F(1,159) = 60.8 $& $<0.0001$ \\
        XLM-R & $F(1,126) = 21.2 $& $<0.0001 $\\
        GPT-2    & $F(1,157) = 64.0 $&$ <0.0001 $\\
        GPT-Neo     & $F(1,152) = 64.1 $&$ <0.0001 $\\
        GPT-J       & $F(1,149) = 62.5 $&$ <0.0001 $\\
        XGLM        & $F(1,146) = 72.6 $ &$ <0.0001$ \\ \hline
        \end{tabular}
    \caption{The results of a Type III ANOVA (using Satterthwaite's method for estimating degrees of freedom; \citealp{kuznetsova_2017_LmerTestPackageTests}) on the \citet{ito_2016_PredictingFormMeaning} stimuli, testing for which language models experimental condition (related or unrelated) is a significant predictor of their surprisal. This is the case for all language models.}
    \label{tab:ito}
    
\end{table}

\subsection{Results}
The results of the experiment are shown in \autoref{fig:ito}. As can be seen, numerically, related words elicit lower surprisals than unrelated words, indicating that they were more highly predicted by the language models. This in turn suggests that these models do in fact collaterally predict the related continuations.

In order to test this more directly, we ran statistical analyses of the surprisals elicited by the language models. This was done by constructing linear mixed-effects regressions for each language model surprisal with experimental condition as a main effect, and the maximal random effects structure that would successfully converge for all models (see \citealp{barr_2013_RandomEffectsStructure}). For all regressions except for that predicting RoBERTa surprisal, this random effects structure was a random intercept of sentence frame and of critical word. For the RoBERTa surprisal regression, the latter random intercept was removed due to it causing a singular fit. As creating null models with only the random effects structure resulted in singular fits for multiple regressions, we were unable to run likelihood ratio tests to test whether experimental condition---that is, whether the word was semantically related or unrelated to the highest-cloze continuation---was a significant predictor of surprisal. For this reason, we instead tested whether experimental condition was a significant predictor of surprisal by running a Type III ANOVA using Satterthwaite's method for estimating degrees of freedom \citep{kuznetsova_2017_LmerTestPackageTests} on the aforementioned linear mixed-effects models that included experimental condition as a fixed effect.

The results of the tests are shown in \autoref{tab:ito}. As can be seen, condition is a significant predictor of the surprisal from every language model, confirming that language models predict related stimuli to be more likely than unrelated stimuli.

\begin{figure*}
    \centering
    \includegraphics[width=\textwidth]{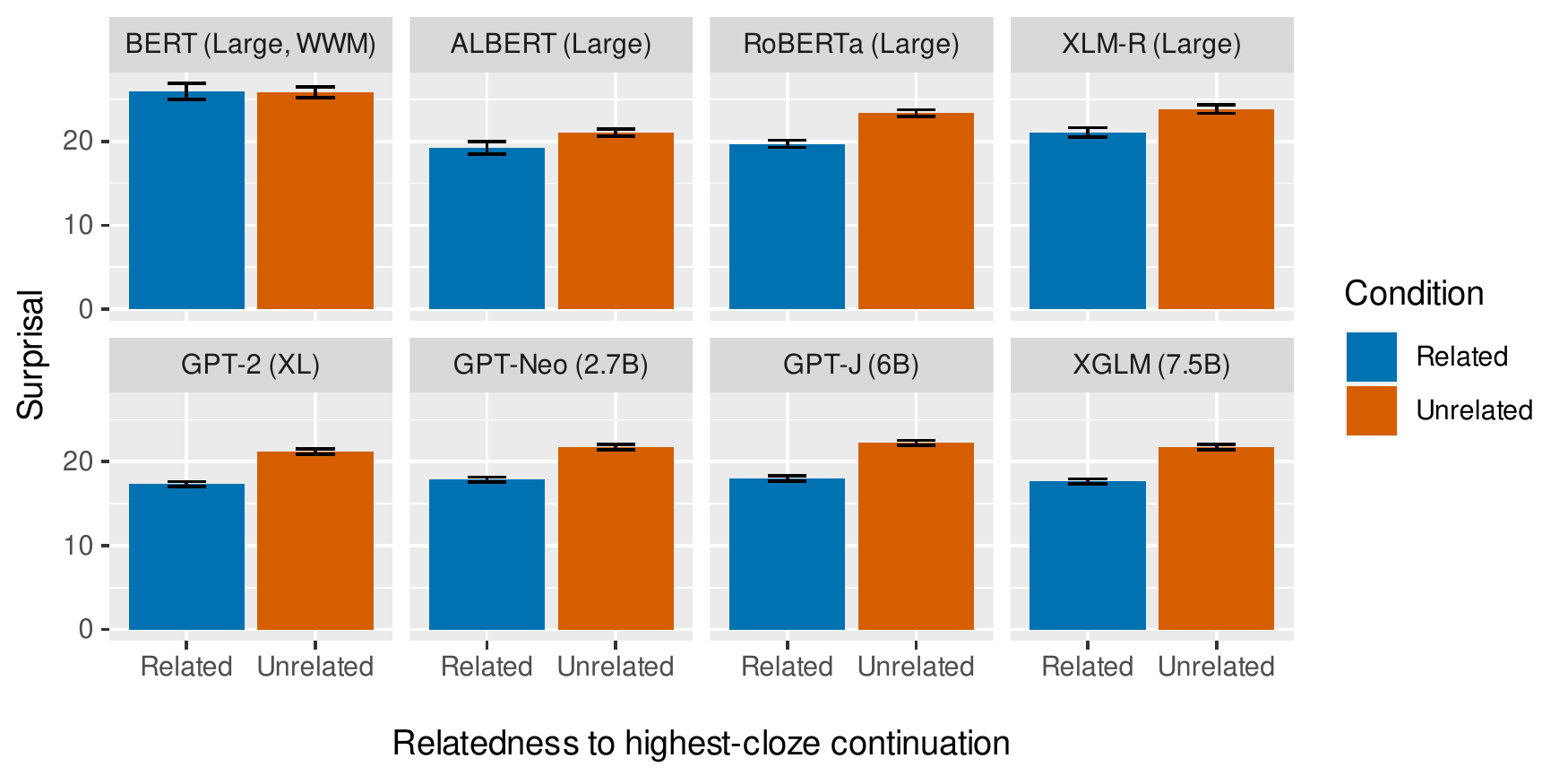}
    \caption{Mean surprisal elicited by each language model for the 
    \citet{delong_2019_SimilarTimeCourses} stimuli related and unrelated to the most probable (highest-cloze) continuation of each sentence. Error bars indicate standard error.}
    \label{fig:delong}
\end{figure*}

The results of this experiment demonstrate that all the language models tested---BERT, ALBERT, RoBERTa, XLM-R, GPT-2, GPT-Neo, GPT-J, and XGLM---display the related anomaly effect in response to the \citet{ito_2016_PredictingFormMeaning} stimuli. All eight models predict implausible continuations that are related to the most probable continuations to be more likely those that are unrelated.

\section{Experiment 2: \citet{delong_2019_SimilarTimeCourses}}

\subsection{Introduction}
\citet{delong_2019_SimilarTimeCourses} also investigated the difference between the N400 amplitude elicited by implausible words that are related or unrelated to the most predictable (highest-cloze) continuation. As in \citet{ito_2016_PredictingFormMeaning}, these stimuli were chosen such that both related and unrelated words were highly implausbile---in this case, `unpredictable words were strategically chosen not to make sense in their given contexts' \citep[p.~4]{delong_2019_SimilarTimeCourses}. These stimuli are exemplified by the set shown in (\ref{ex:delong_example}).

\begin{example}
\label{ex:delong_example}
The commuter drove to work in her \_\_\_\_\_\_ after breakfast.
\begin{itemize}
\item \textit{car} (Predictable)
\item \textit{brakes} (Related)
\item \textit{poetry} (Unrelated)
\end{itemize}
\end{example}

Like \citet{ito_2016_PredictingFormMeaning}, \citet{delong_2019_SimilarTimeCourses} find that overall, related continuations elicit a smaller N400 response than unrelated continuations.

\begin{figure*}[bp]
    \centering
    \includegraphics[width=\textwidth]{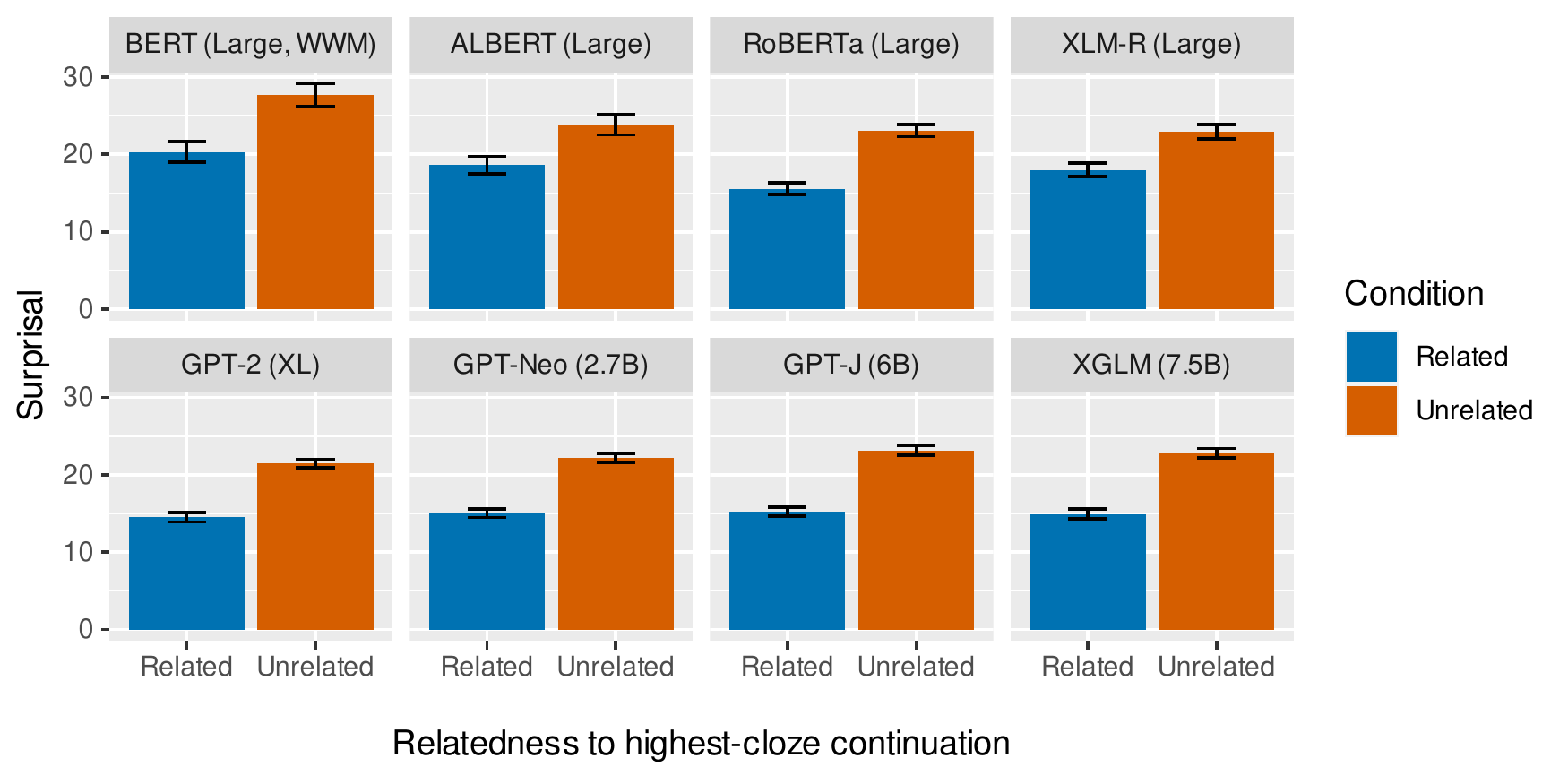}
    \caption{Mean surprisal elicited by each language model for the 
    \citet{metusalem_2012_GeneralizedEventKnowledge} stimuli related and unrelated to the most probable (highest-cloze) continuation of each sentence. Error bars indicate standard error.}
    \label{fig:metusalem}
\end{figure*}

\begin{table}[]
\renewcommand{\arraystretch}{1.2}
    \centering
        \begin{tabular}{llr}
        \hline
        \textbf{Model}    & \textbf{Test Statistic}   & \textbf{Corrected \textit{p}}       \\ \hline
        BERT     & $F(1,159) = <0.1  $&$ 0.9322            $\\
        ALBERT   & $F(1,112) = 6.3  $&$ 0.0138          $\\
        RoBERTa  & $F(1,159) = 50.7 $& $<0.0001           $\\
        XLM-R    & $F(1,132) = 18.2   $&$ 0.0001           $\\
        GPT-2 XL & $F(1,134) = 120.7  $&$ <0.0001           $\\
        GPT-Neo  & $F(1,142) = 111.7 $&$ <0.0001           $\\
        GPT-J    & $F(1,141) = 132.6 $&$ <0.0001 $\\
        XGLM     & $F(1,159) = 122.4   $&$ <0.0001           $\\ \hline
        \end{tabular}
    \caption{The results of a Type III ANOVA (using Satterthwaite's method for estimating degrees of freedom; \citealp{kuznetsova_2017_LmerTestPackageTests}) on the \citet{delong_2019_SimilarTimeCourses} stimuli, testing for which language models experimental condition (related or unrelated) is a significant predictor of their surprisal. This is the case for all language models except BERT.}
    \label{tab:delong}
\end{table}

\subsection{Results}

As in Experiment 1, we ran the stimuli from the original experiment through the 8 language models and calculated the surprisal of each critical word. The results of the experiment are shown in \autoref{fig:delong}. In all models except BERT, related stimuli all elicit numerically lower surprisals than unrelated stimuli, indicating that they were more highly-predicted by the language models.

We again ran the same statistical test as in Experiment 1, testing whether experimental condition (related or unrelated to the highest-cloze continuation) is a significant predictor of the surprisal elicited by the stimuli in each language model. The ALBERT, XLM-R, GPT-2, GPT-Neo, and GPT-J regressions had random intercepts of sentence frame and critical word, while the BERT, RoBERTa, and XGLM regressions had only random intercepts for sentence frame. The results of the Type III ANOVA are shown in \autoref{tab:delong}. Condition is a significant predictor of the surprisal of every model except BERT---in these models, related stimuli are predicted to be more likely continuations of the sentence than unrelated stimuli. Thus, with the exception of BERT, we replicate the findings of Experiment 1. 

\section{Experiment 3: \citet{metusalem_2012_GeneralizedEventKnowledge} }

\subsection{Introduction}
 \citet{metusalem_2012_GeneralizedEventKnowledge} investigated the extent to which relatedness to the event described in the preceding context impacts the amplitude of the N400 response. \citet{metusalem_2012_GeneralizedEventKnowledge} presented human participants with experimental stimuli that included either the most probable (highest-cloze) continuation of a sentence, an implausible continuation that was related to the event described, or an implausible continuation that was unrelated to the event described. All of the implausible stimuli also had a cloze probability of zero. The stimuli are exemplified by the set for a single sentence frame shown in (\ref{ex:metusalem_example2}).

\begin{example}
\label{ex:metusalem_example2}
We're lucky to live in a town with such a great art museum. Last week I went to see a special exhibit. I ﬁnally got in after waiting in a long \_\_\_\_\_\_.
\begin{itemize}
\item \textit{line} (Predictable)
\item \textit{painting} (Related)
\item \textit{toothbrush} (Unrelated)
\end{itemize}
\end{example}

\citet{metusalem_2012_GeneralizedEventKnowledge} found that despite their implausibility and improbability (based on cloze), critical words related to the event described in the context preceding them elicited smaller N400 responses than words that were unrelated to the event, a clear example of a related anomaly effect.

\subsection{Results}

As in Experiments 1 and 2, we ran the stimuli from the original experiment through the 8 language models and calculated the surprisal of each critical word. The results of the experiment are shown in \autoref{fig:metusalem}. As in Experiment 1, numerically, in all models related stimuli elicit lower surprisals than unrelated surprisals, indicating that they were more highly predicted by the language models.

We again ran the same statistical analyses as in Experiments 1 and 2, constructing linear mixed-effects regression models, all of which had random intercepts of sentence frame and critical word. Using a Type III ANOVA, we tested whether experimental condition (related or unrelated to the event described in the preceding context) is a significant predictor of N400 amplitude. The results are shown in \autoref{tab:metusalem}. As can be seen, experimental condition was a significant predictor of the surprisal of all models.

\begin{table}[]
\renewcommand{\arraystretch}{1.2}
    \centering
        \begin{tabular}{llr}
        \hline
        \textbf{Model}    & \textbf{Test Statistic}    & \textbf{Corrected \textit{p}}       \\ \hline
        BERT     & $F(1,29) = 77.1  $& $<0.0001 $\\
        ALBERT   & $F(1,29) = 78.7  $& $<0.0001 $\\
        RoBERTa  & $F(1,28) = 188.1  $& $<0.0001 $\\
        XLM-R    & $F(1,34) = 83.4  $& $<0.0001 $\\
        GPT-2 XL & $F(1,35) = 211.5 $& $<0.0001 $\\
        GPT-Neo  & $F(1,42) = 200.1 $& $<0.0001 $\\
        GPT-J    & $F(1,35) = 265.5 $& $<0.0001 $\\
        XGLM     & $F(1,33) = 222.5 $& $<0.0001 $\\ \hline
        \end{tabular}
    \caption{The results of a Type III ANOVA (using Satterthwaite's method for estimating degrees of freedom; \citealp{kuznetsova_2017_LmerTestPackageTests}) on the \citet{metusalem_2012_GeneralizedEventKnowledge} stimuli, testing for which language models experimental condition (related or unrelated) is a significant predictor of their surprisal. This is the case for all language models.}
    \label{tab:metusalem}
\end{table}

\section{General Discussion}
\subsection{Summary of Results}
In all but one specific case---BERT in Experiment 2---experimental condition significantly predicted language model surprisal in the same direction as human N400 responses. The results of Experiments 1 and 2, therefore demonstrate convincingly that, like humans, language models do tend to predict that anomalous words related to the most probable continuation are more probable than anomalous words that are not. The results of Experiments 3, analogously, demonstrate that like humans, language models tend to predict that anomalous words related to a relevant event described in the preceding context are more probable than anomalous words that are not. Thus, like the human language comprehension system, language models exhibit related anomaly effects.

\subsection{Psycholinguistic implications}
These results have clear implications for psycholinguistic research on the effects of related anomalies on human language processing. First, a predictive system can display the effects---in fact, there is only one set of stimuli for which not all models do. This demonstrates the sufficiency of a predictive system for preactivating related anomalous stimuli to a greater degree than unrelated anomalous stimuli. In other words, based on a parsimony criterion, there is no need to posit that related anomaly effects on human language processing require something beyond a predictive system such as an associative system, either instead of or in addition to a predictive one.

Second, both kinds of related anomaly effect explored---the reduction in N400 amplitude correlated with relatedness to the most probable continuation and that correlated with relatedness to the event in the preceding context---are explainable by a single mechanism. This may seem counterintuitive, given how intuitively different the effects may seem. Yet this finding is consistent with the idea in the literature that the two effects can be considered different variants of the same phenomenon \citep{delong_2019_SimilarTimeCourses,delong_2020_ComprehendingSurprisingSentences}.

Given that this study is based on computational modeling, we should note that the results do not constitute direct proof of a neurocognitive predictive system or of the lack of the involvement of an additional associative mechanism. However, they are consistent with such accounts, and open the door for future research, both computational and experimental. For example, it may be the case that other phenomena that have been argued to constitute evidence for a separate associative mechanism \citep[see][for review]{federmeier_2021_ConnectingConsideringElectrophysiology} may also be explainable on the basis of prediction. On the other hand, the approach we use here can also be used to design stimuli that do not differ in probability in order to further test whether prediction can explain all related anomaly effects.

\subsection{Implications for NLP}
The results of the present study demonstrate that related anomaly effects occur in contemporary transformer language models. Based on the present study, this does not appear to be impacted by whether the model is an autoregressive or masked language model; or by whether the model is monolingual or multilingual. In fact, the only model that does not show the effect every time is BERT, the least powerful model tested (all other models are either larger, trained on more data, or both). Thus, in line with previous research showing that higher-quality language models better predict human processing metrics \citep{merkx_2021_HumanSentenceProcessing}, the present results suggest that better language models are also more likely to display human-like patterns of prediction.

The results of this study also have several implications for understanding how the predictions of humans and language models relate. As has been previously discussed, some researchers have argued that we should evaluate the predictions of language models based on cloze probability \citep{ettinger_2020_WhatBERTNot}. In fact, some have suggested training models on cloze probabilities \citep{eisape_2020_ClozeDistillationImproving}. However, the results of this study, along with others \citep{frank_2015_ERPResponseAmount,aurnhammer_2019_EvaluatingInformationtheoreticMeasures,michaelov_2020_HowWellDoes,aurnhammer_2019_EvaluatingInformationtheoreticMeasures,merkx_2021_HumanSentenceProcessing,szewczyk_2022_ContextbasedFacilitationSemantic,michaelov_2022_ClozeFarN400}, suggest that the predictions of language models are highly correlated with N400 amplitude; and recent work has argued that that the activation states of transformers are highly correlated with activation in the brain during language comprehension more generally \citep{schrimpf_2020_NeuralArchitectureLanguage}. Thus, while it may be useful for certain tasks to have cloze-like predictions, it may be the case that we are generally more likely to get N400-like predictions from language models.

If so, this is a cause for both optimism and pessimism. Given that humans are the gold-standard in natural language tasks generally, if a language model can make predictions that closely match those that humans make as part of language comprehension, this may also suggest that the representations learned are at least in some ways functionally similar to those that humans use to generate the same predictions. On the other hand, by the same token, it may suggest a limit to the possibilities of language modeling alone---there is much more to language comprehension than the kinds of prediction that underlie the N400 response \citep[see, e.g.,][]{ferreira_2019_ProblemComprehensionPsycholinguistics,delong_2020_ComprehendingSurprisingSentences,kuperberg_2020_TaleTwoPositivities}.

\section{Conclusion}
In order to better understand related anomaly effects in humans, we investigated whether contemporary transformer language models display them. We found that in all but one case, they do, suggesting that related anomaly effects in both humans and language models may be driven by prediction alone.

\section*{Acknowledgements}
We would like to thank the authors of the original N400 experiment papers---Wen-Hsuan Chan, 
Martin Corley, 
Katherine A. DeLong, 
Jeffrey L. Elman, 
Mary Hare, 
Aine Ito, 
Marta Kutas, 
Andrea E. Martin, 
Ken McRae, 
Ross Metusalem, 
Mante S. Nieuwland,
Martin J. Pickering, and
Thomas P. Urbach---for making their stimuli available. We would also like to thank the anonymous reviewers for their helpful comments, the other members of the Language and Cognition Lab at UCSD for their valuable discussion, and the San Diego Social Sciences Computing Facility Team for their technical assistance. This work was partially supported by a 2021-2022 Center for Academic Research and Training in Anthropogeny Annette Merle-Smith Fellowship awarded to James A. Michaelov, and the RTX A5000 used for this research was donated by the NVIDIA Corporation.

\bibliography{anthology,custom,library}
\bibliographystyle{acl_natbib}

\appendix

\section{Limitations}
As mentioned the discussion section, one limitation of the present study is that while it demonstrates that it is possible for related anomaly effects to emerge from a system engaged in prediction alone, it does not directly demonstrate that this is what is occurring in humans.

A further limitation is that we model the results of three related anomaly experiments out of the larger total number that have been carried out \citep[for review, see][]{delong_2019_SimilarTimeCourses}. However, given how consistent related anomaly effects appear to be \citep{delong_2019_SimilarTimeCourses}, and how consistent our results are (after statistical correction for multiple comparisons, all three related anomaly effects are modeled by all but one transformer, which only fails to model one effect), we do not believe this presents a problem for our analysis.

Finally, the three experiments modeled were all carried out in English. Related anomaly effects have been reported in other languages \citep{delong_2019_SimilarTimeCourses} such as Dutch \citep{rommers_2013_ContentsPredictionsSentence}; and these are not modeled in our study. Thus, it is an open question whether our results generalize to related anomaly effects in languages other than English. However, we also note the evidence that higher-quality models are better at predicting N400 amplitude \citep{merkx_2021_HumanSentenceProcessing}. For this reason, given the overwhelming focus on English in computational linguistics \citep{bender_2009_LinguisticallyNaiveLanguage,bender_2011_AchievingEvaluatingLanguageIndependence,tsarfaty_2013_ParsingMorphologicallyRich,munro_2015_LanguagesACLThis,mielke_2016_LanguageDiversityACL,kim_2016_CharacterAwareNeuralLanguage,amram_2018_RepresentationsArchitecturesNeural,bender_2019_BenderRuleNamingLanguages,clark_2022_CaninePretrainingEfficient}, current language model architectures are likely to be best suited to predicting English---indeed, current state-of-the-art models such as GPT-3 \citep{brown_2020_LanguageModelsAre}, OPT \citep{zhang_2022_OPTOpenPretrained},  PaLM \citep{chowdhery_2022_PaLMScalingLanguage}, and LaMDA \citep{thoppilan_2022_LaMDALanguageModels} are trained mostly or only on English data. Thus, while the focus on modeling English may be an issue for the field as a whole, in this case, focusing on experiments carried out in English may in fact give us the best possible chance to evaluate what the human predictive system \textit{could} predict.

\section{Models used}
The details of the models used in this study are provided in \autoref{tab:appendix}.

\begin{table*}[hb]
\renewcommand{\arraystretch}{1.2}
    \centering
        \begin{tabular}{lll}
        \hline
        \textbf{Model Name}    & \textbf{Full Name on the Hugging Face Model Hub}    & \textbf{Reference}       \\ \hline
        BERT     & \texttt{bert-large-cased-whole-word-masking}& \citet{devlin_2019_BERTPretrainingDeep}\\
        ALBERT   &  \texttt{albert-xxlarge-v2}& \citet{lan_2020_ALBERTLiteBERT}\\
        RoBERTa  & \texttt{roberta-large}& \citet{liu_2019_RoBERTaRobustlyOptimized}\\
        XLM-R    & \texttt{xlm-roberta-large}& \citet{conneau_2020_UnsupervisedCrosslingualRepresentation}\\
        GPT-2 XL &  \texttt{gpt2-xl}& \citet{radford_2019_LanguageModelsAre}\\
        GPT-Neo  &  \texttt{EleutherAI/gpt-neo-2.7B}& \citet{black_2021_GPTNeoLargeScale}\\
        GPT-J    &  \texttt{EleutherAI/gpt-j-6B}& \citet{wang_2021_GPTJ6BBillionParameter}\\
        XGLM     &  \texttt{facebook/xglm-7.5B}& \citet{lin_2021_FewshotLearningMultilingual}\\ \hline
        \end{tabular}
        
    \caption{Transformer langauge models used in the present study. All were accessed using the transformers \citep{wolf_2020_TransformersStateoftheArtNatural} package.}
    \label{tab:appendix}
\end{table*}

\end{document}